\def\year{2018}\relax
\documentclass[letterpaper]{article} 
\usepackage{aaai18}  
\usepackage{times}  
\usepackage{helvet}  
\usepackage{courier}  
\usepackage{url}  
\usepackage{graphicx}  
\usepackage{amsfonts}
\usepackage{amsmath}
\usepackage{amssymb}
\usepackage{subcaption}
\graphicspath{{./figures/}}

\newcommand{\bv}{\mathbf{b}} 
\newcommand{\cv}{\mathbf{c}}

\newcommand{\hv}{\mathbf{h}}

 \newcommand{\Rv}{\mathbf{R}}

 \newcommand{\Vv}{\mathbf{V}}
 \newcommand{\Wv}{\mathbf{W}}

\newcommand{\thetav}{\boldsymbol{\theta}}

\newcommand{\ep}{\mathop{\mathbb{E}}}

\newcommand{\data}{\mathcal{D}}
\newcommand{\indicator}{\mathbb{I}}

\newcommand{\RR}{\mathbb{R}}

\newcommand{\fig}[1]{Fig.~\ref{fig:#1}}
\newcommand{\tabl}[1]{Table~\ref{table:#1}}
\newcommand{\eqn}[1]{Eqn.~(\ref{eqn:#1})}
\newcommand{\secref}[1]{Sec.~\ref{sec:#1}}
\newcommand{\citet}[1]{\citeauthor{#1}~\shortcite{#1}}

\newcommand{\method}{CF-UIcA}
\newcommand{\mlone}{MovieLens 1M}
\newcommand{\netflix}{Netflix}

\newcommand{\Rij}{R_{i,j}}

\newcommand{\Rcondu}{R_{o_{t}}^{IIC}}
\newcommand{\Rcondi}{R_{o_{t}}^{UUC}}

\begin{document}
%
\title{Collaborative Filtering with User-Item Co-Autoregressive Models}
\author{Chao Du$^{\dagger}$ \quad Chongxuan Li$^{\dagger}$ \quad Yin Zheng$^{\ddagger}$ \quad Jun Zhu\thanks{corresponding author.}$^{\dagger}$ \quad Bo Zhang$^{\dagger}$\\
$^{\dagger}$Dept. of Comp. Sci. \& Tech., State Key Lab of Intell. Tech. \& Sys., TNList Lab,\\
$^{\dagger}$Center for Bio-Inspired Computing Research, Tsinghua University, Beijing, 100084, China\\
$^{\ddagger}$Tencent AI Lab, Shenzhen, Guangdong, China\\
}
\maketitle
\begin{abstract}
Deep neural networks have shown promise in collaborative filtering (CF). However, existing neural approaches are either user-based or item-based, which cannot leverage all the underlying information explicitly. We propose CF-UIcA, a neural co-autoregressive model for CF tasks, which exploits the structural correlation in the domains of both users and items. The co-autoregression allows extra desired properties to be incorporated for different tasks. Furthermore, we develop an efficient stochastic learning algorithm to handle large scale datasets. We evaluate CF-UIcA on two popular benchmarks: MovieLens 1M and Netflix, and achieve state-of-the-art performance in both rating prediction and top-N recommendation tasks, which demonstrates the effectiveness of CF-UIcA.
\end{abstract}

\section{Introduction}
With the fast development of electronic commerce, social networks and music/movie content providers, 
recommendation systems have attracted extensive research attention~\cite{burke2002hybrid,schafer2007collaborative}.
As one of the most popular methods,
collaborative filtering (CF)~\cite{schafer2007collaborative,billsus1998learning,resnick1994grouplens,salakhutdinov2007restricted} predicts users' preferences for items based on their previous behaviors (rating/clicking/purchasing etc.) in a recommendation system.
CF enjoys the benefit of content-independence of the items being recommended. Thus, it does not need expert knowledge about the items when compared with content-based methods~\cite{van2013deep,gopalan2014content} and could possibly provide cross-domain recommendations.

The basic assumption behind CF is that
there exist correlations between the observed user behaviors, and these correlations can be generalized to their future behaviors.
Basically, the correlations can be categorized as \textit{User-User Correlations} (UUCs)---the correlations between different users' behaviors on a same item,
and \textit{Item-Item Correlations} (IICs)---the correlations between 
a user's behaviors on different items.
These two types of underlying correlations usually exist crisscrossing in the partially observed user behaviors, making CF a difficult task.

Extensive work has studied how to effectively exploit the underlying
correlations to make accurate predictions. Early approaches~\cite{resnick1994grouplens,sarwar2001item} consider  UUCs or IICs by
computing the similarities between users or items.
As one of the most popular classes of CF methods,
matrix factorization (MF)~\cite{billsus1998learning,koren2009matrix,salakhutdinov2007probabilistic} assumes that the partially observed matrix (of ratings) is low-rank and embeds both users and items into a shared latent space.
MF methods consider both UUCs and IICs implicitly as a prediction is simply the inner product of the latent vectors of the corresponding user and item.
Recently, deep learning methods have achieved promising results in various tasks~\cite{bahdanau2014neural,mnih2015human,silver2016mastering} due to their ability to learn a rich set of abstract representations. 
Inspired by these advances, neural networks based CF methods~\cite{salakhutdinov2007restricted,sedhain2015autorec,wu2016collaborative,zheng2016neural}, which employ highly flexible transformations to model a user's behavior profile (all behaviors) with a compact representation,
are widely studied
as alternatives to MF.
These methods essentially consider all UUCs explicitly as they take inputs of users' all observed behaviors.
(See more details in \secref{basic}.)

\begin{figure*}[tb]
\centering
\begin{subfigure}[b]{.24\linewidth}
\includegraphics[width=0.95\textwidth]{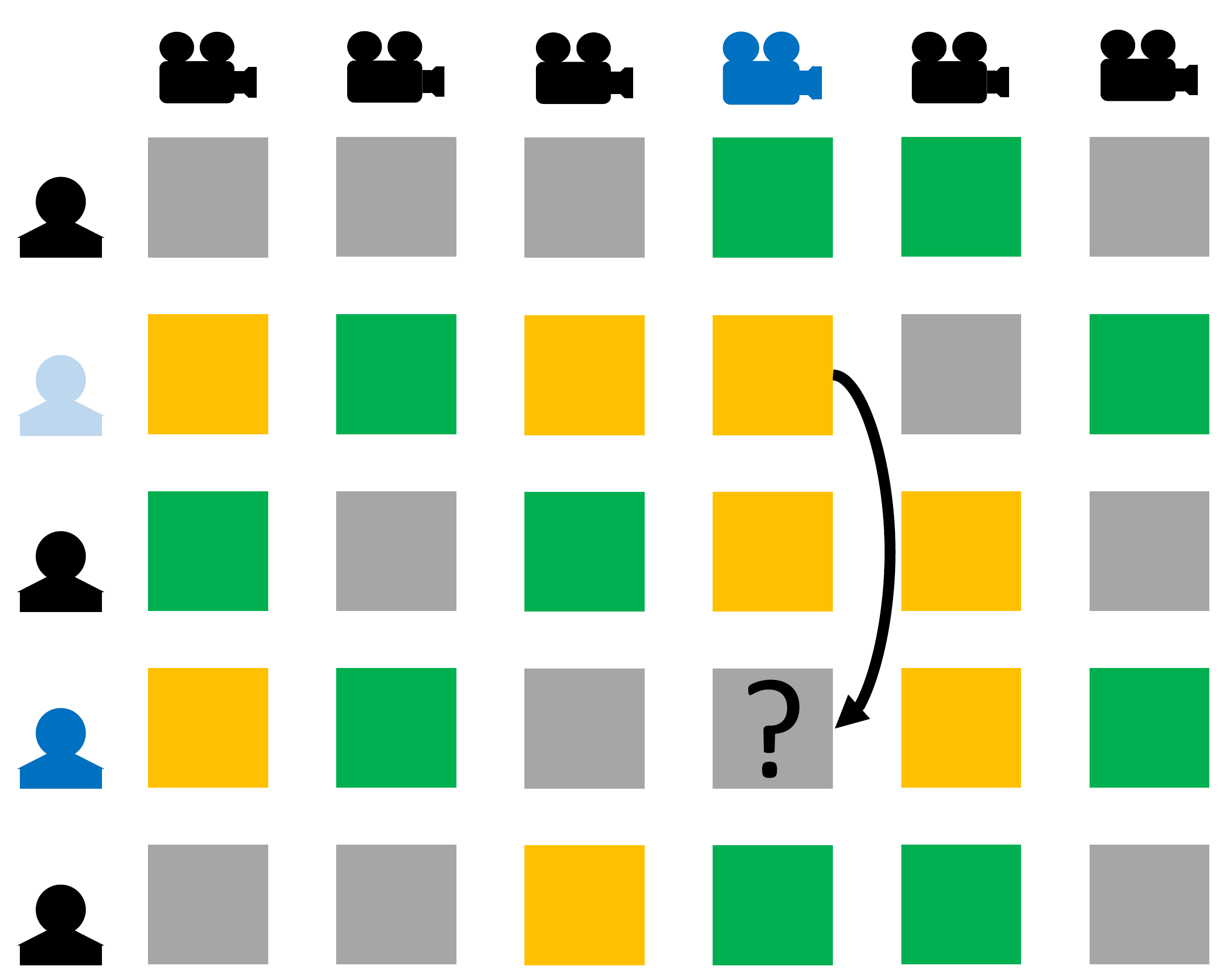}
\caption{}\label{fig:basica}
\end{subfigure}
\hfill
\begin{subfigure}[b]{.24\linewidth}
\includegraphics[width=0.95\textwidth]{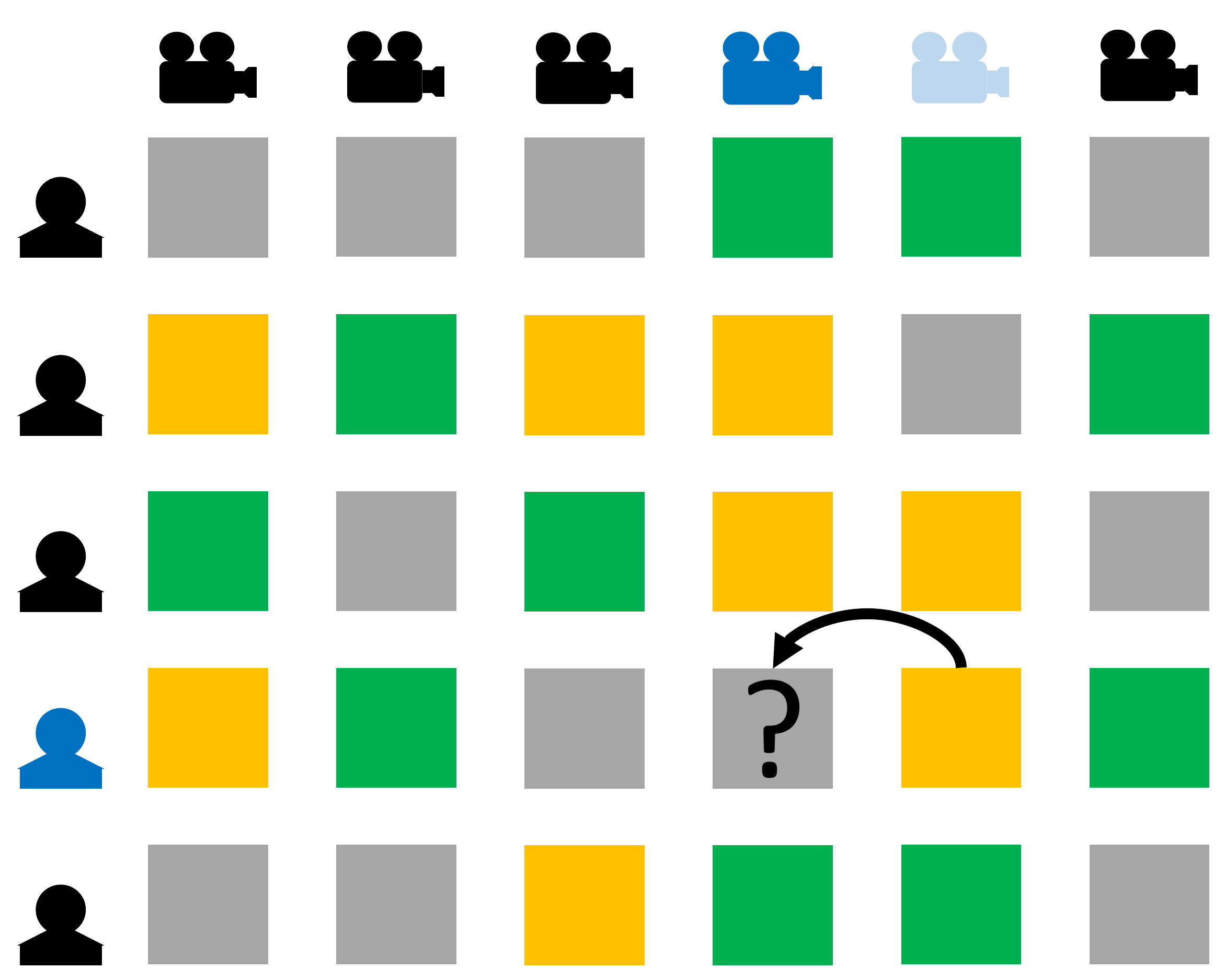}
\caption{}\label{fig:basicb}
\end{subfigure}
\hfill
\begin{subfigure}[b]{.24\linewidth}
\includegraphics[width=0.95\textwidth]{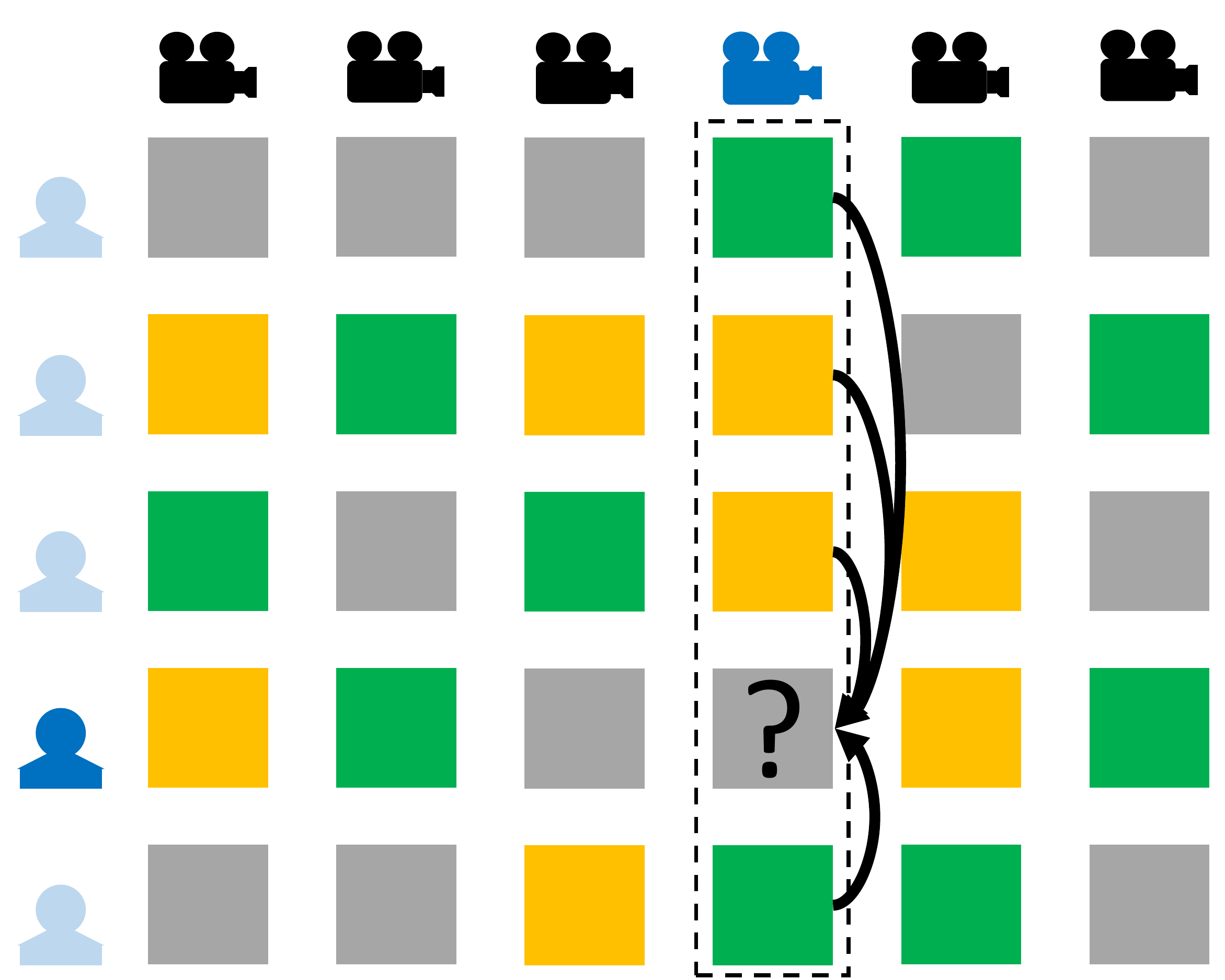}
\caption{}\label{fig:basicc}
\end{subfigure}
\hfill
\begin{subfigure}[b]{.24\linewidth}
\includegraphics[width=0.95\textwidth]{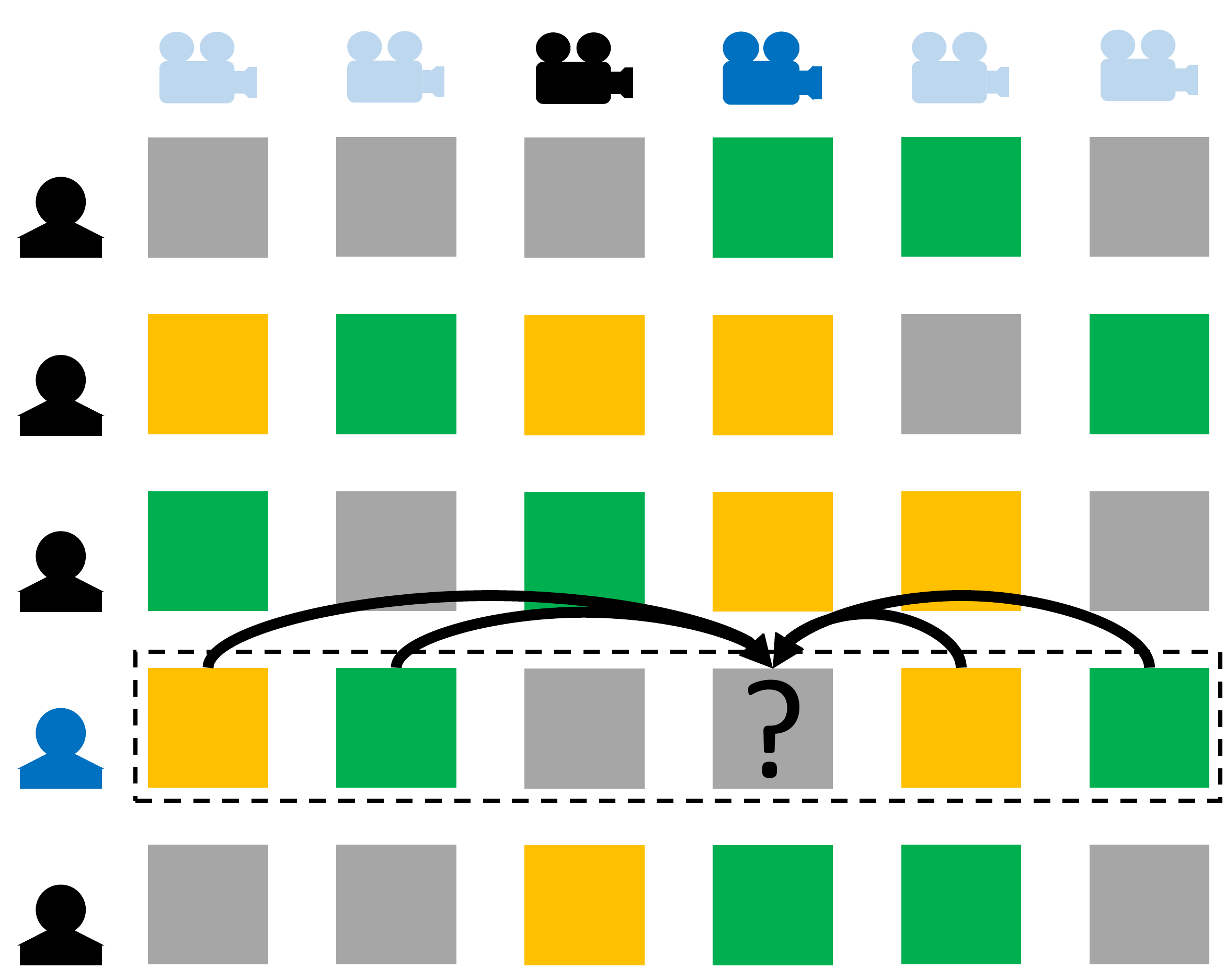}
\caption{}\label{fig:basicd}
\end{subfigure}
\caption{Illustration of predictions in a toy recommendation system with 5 users and 6 items (best viewed in color). Each square in green/yellow/gray corresponds to a positive/negative/unobserved behavior, respectively. The behavior (whose associating user and item are colored in deep blue) being predicted is marked with a question mark.
(a) Predict with a single User-User Correlation: the behavior is predicted according to the behavior of another user (labeled as light blue);
(b) Predict with a single Item-Item Correlation; (c) Predict with multiple User-User Correlations, and (d) Predict with multiple Item-Item Correlations.}\label{fig:basic}
\end{figure*}

Though previous neural networks based methods are promising,
one common drawback of these methods lies in that they cannot exploit both UUCs and IICs together,
making them further improvable.
%
To this end, 
we propose a novel neural autoregressive model for CF,
named User-Item co-Autoregressive model (\method{}), which considers the autoregression in the domains of both users and items,
so as to model both UUCs and IICs together.
The introduced co-autoregression naturally provides a principled way to select the UUCs and IICs to be generalized that are helpful for prediction.
This allows us to incorporate extra desired properties into the model for different tasks,
which is not studied in previous work.
We further develop a stochastic learning algorithm for \method{} to make it applicable for large datasets.
We demonstrate our method on two real benchmarks: \mlone{} and \netflix{},
achieving state-of-the-art results in both rating prediction and top-N recommendation tasks, which is rarely accomplished in previous work.
In addition, the
visualization demonstrates that \method{} learns semantic patterns without extra supervision.

\section{Related Work}\label{sec:basic}

Collaborative filtering methods make predictions based on user behaviors,
which could reveal certain patterns for generalization.
These phenomena involve two types of information: \textit{User-User Correlations} (UUCs) and 
\textit{Item-Item Correlations} (IICs).
As shown in~\fig{basica}, UUC depicts that a user's behavior
is usually related to the one of some other users on the same item, especially when they have similar habits/tastes.
Similarly,
IIC depicts that a user's behavior on an item is related to his/her behaviors on other items, especially when these items are similar in nature, as shown in \fig{basicb}.
Predictions are then possible to be made by integrating these correlations.
\fig{basicc} and \fig{basicd} show all the UUCs and IICs of the unknown preference marked by the question mark.
Intuitively, integrating multiple UUCs and IICs can potentially lead to a more precise prediction.

Existing CF methods either implicitly or explicitly exploit these correlations. Early methods model the correlations via some similarity functions on the raw preferences,
such as k-NN collaborative filtering (kNN-CF)~\cite{resnick1994grouplens,sarwar2001item}.
These methods make predictions with the top $k$ UUCs or IICs explicitly. 
Matrix factorization (MF) methods characterize both users and items by vectors in a low-dimensional latent space.
The predictions are modeled with the inner products of the latent vectors of the corresponding users and items. Representative works include SVD-based methods~\cite{billsus1998learning,sarwar00applicationof} and the probabilistic MF (PMF)~\cite{salakhutdinov2007probabilistic,salakhutdinov2008bayesian}.
Recent approaches improve MF by loosing the constrains of linearity and low-rank assumption.
Bias MF~\cite{koren2009matrix} introduces bias terms associated with users and items.
\citet{lee2013local} propose Local Low-Rank Matrix Approximation (LLORMA) by assuming the observed rating matrix is a weighted sum of low-rank matrices.
NNMF~\cite{dziugaite2015neural} and NeuMF~\cite{he2017neural} replace the inner product operations in MF with neural networks.
Since MF methods make predictions with the learned latent vectors of the users and the items, the UUCs and IICs are not modeled explicitly.

With the success in
many tasks~\cite{krizhevsky2012imagenet,graves2013speech,bahdanau2014neural,mnih2015human,silver2016mastering}, deep learning has been integrated into CF methods with great success.
~\citet{salakhutdinov2007restricted} propose RBM-CF, a CF methods based on Restricted Boltzmann Machines, which has shown its power in Netflix prize challenge~\cite{bennett2007netflix}.
Sedhain et al.~\shortcite{sedhain2015autorec} propose AutoRec, a discriminative model based on auto-encoders.
A similar model known as CDAE is concurrently proposed by~\citet{wu2016collaborative}.
Recently, Zheng et al.~\shortcite{zheng2016neural} propose CF-NADE, a tractable model based on
Neural Autoregressive Distribution Estimators (NADE)~\cite{larochelle2011neural},
and achieve the state-of-the-art results on several CF benchmarks. 
These methods share two aspects: 1) different models are built for different users by sharing parameters; and 2) predictions are made for a user according to
his/her behavior profile.
Note that as the role of users and items are exchangeable, these methods usually have a user-based and an item-based variants. As a result, these methods make predictions with either the UUCs or the IICs explicitly.

Our \method{} differs from existing CF methods in that it can capture both UUCs and IICs explicitly and simultaneously.
Similar as in CF-NADE, we adopt neural autoregressive architectures to model the probabilities of the behaviors.
The crucial difference is that CF-NADE models the rating vectors of each user, making the users independent from each other, while
\method{} models the behaviors across all users and items in order to consider UUCs and IICs jointly.
Moreover, we analyze the significance of the co-autoregression in a novel perspective  and demonstrate its effectiveness, which is another step beyond CF-NADE.

\begin{figure*}[tb]
\centering
\includegraphics[width=0.85\textwidth,height=0.26\textwidth]{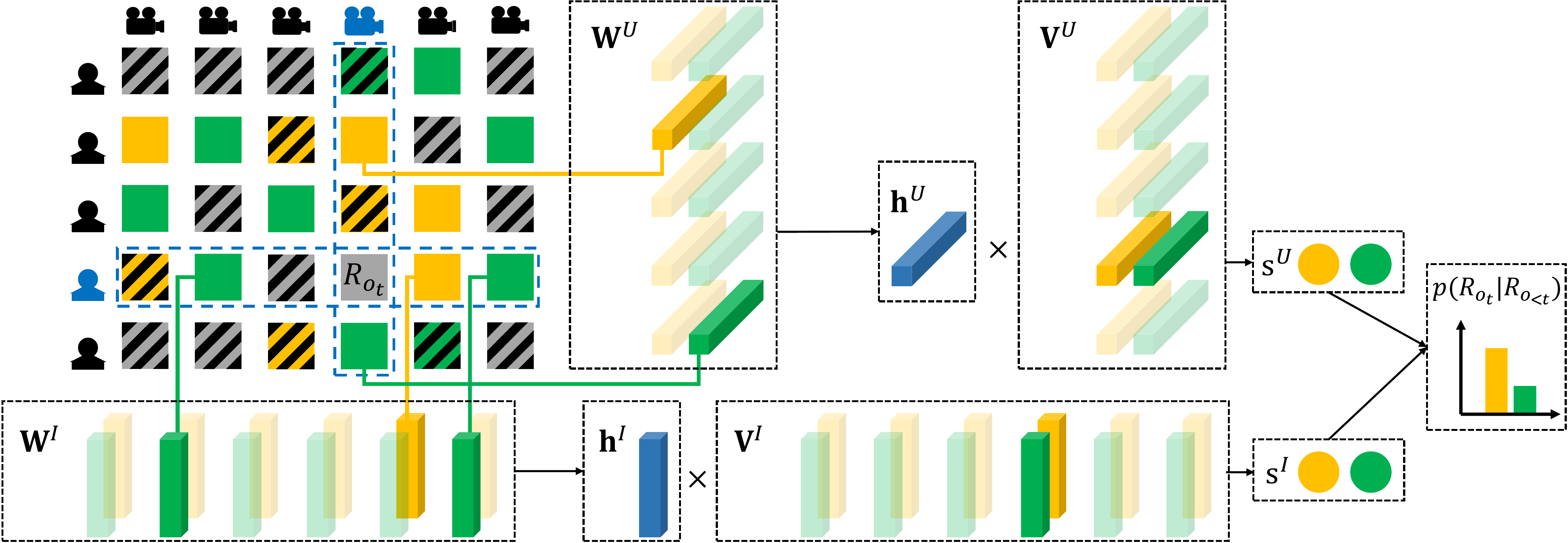}
\caption{An illustration of the conditional model. The yellow, green and gray entries are interpreted same as in~\fig{basic}. Suppose $R_{o_t}$ is the current behavior being modeled. The black striped lines mark the entries of $R_{o_{>t}}$. The blue dashed boxes line out the UUCs and IICs for $R_{o_{t}}$. The cuboids represent the columns of $\Wv^U,\Wv^I,\Vv^U,\Vv^I$, with the color corresponding to the behaviors. The hidden representations $\hv^U$ and $\hv^I$ are computed by summing over the corresponding columns (with the uncorresponding columns marked in lighter colors) of UUCs and IICs.
The activations $s^U$ and $s^I$ are computed by multiplying the hidden representations and the corresponding columns of $\Vv^U$ and $\Vv^I$.
We omit the bias terms for clarity.}\label{fig:workflow}
\end{figure*}

Hybrid recommendation \cite{burke2002hybrid} is a class of methods focusing on combining different techniques at a high level, e.g., combining CF-based methods and content-based methods together. Different with the existing hybrid recommendation methods, our model focuses on utilizing both user-based and item-based information.
\citet{wang2006unifying} share similar motivation with ours. However, their method is memory-based and unifies user-based and item-based models by similarity. While our method is model-based and combines user-based and item-based information by autoregressive neural networks.

\section{Method}\label{sec:model}

We now present \method{} which models both UUCs and IICs with co-autoregressvie architectures.

Let $N$ and $M$ denote the number of users and items, respectively.
We define the behavior matrix $\Rv\in\RR^{N\times M}$ from user behaviors by assigning entries of $\Rv$ with different labels for different behaviors:
For explicit feedback, e.g. $K$-star scale ratings, we define $\Rij=k$ for the behavior ``user $i$ rates item $j$ with $k$ stars''; For implicit feedback, e.g. clicks, we define $\Rij=1$ for the behavior ``user $i$ clicks item $j$''. And we define $\Rij=0$ for unobserved entries.
Let $\data=\{R_{i_d,j_d}\}_{d=1}^D$ be all the observed behaviors, which form the training set.
The goal of CF is then to predict an unknown behavior $R_{i^*,j^*}\not\in\data$ based on the observed behaviors $\data$.

\subsection{The Model}

Autoregressive models~\cite{frey1998graphical,larochelle2011neural,JMLR:v18:16-017} offer a natural way to introduce interdependencies, which is desired for CF tasks, as analyzed in \secref{basic}.
We start with a very generic autoregressive assumption to model the probability of the behavior matrix:
{\small
\begin{equation}\label{eqn:prob-R}
    p(\Rv) = \prod_{t=1}^{N\times M} p( R_{o_t}| R_{o_{<t}}),
\end{equation}}
where $o$ is a permutation of 
all $\langle$user, item$\rangle$ pairs
that serves as an ordering of all the entries in the behavior matrix $\Rv$, and $R_{o_{<t}}$ denotes the first $t\!-\!1$ entries of $\Rv$ indexed by $o$.
For example, $o_t\!=\!(i',j')$ indicates that the behavior $R_{i',j'}$ is at the $t$-th position in $o$, i.e., $R_{o_t}\!=\!R_{i',j'}$.
Let $o_t^i\!=\!i'$ and $o_t^j\!=\!j'$ denote the first and second dimension of $o_t$, which index the user and the item, respectively.

Basically, there are $(N\times M)!$ possible orderings of all the entries of $\Rv$. For now we assume that $o$ is fixed. (We will discuss the orderings latter.) If we consider $o$ as the ordering of the timestamps of the behaviors observed by the system, then the conditional in \eqn{prob-R} means that the model predicts behavior $R_{o_t}$ at time $t$ depends on all the observed behaviors before $t$.

\subsubsection{The Conditional Model}

According to \secref{basic}, both UUCs and IICs are informative for prediction.
We therefore define a conditional model that exploits both UUCs and IICs:
{\small
\begin{equation}\label{eqn:conditional}
    p( R_{o_t}| R_{o_{<t}})=p( R_{o_t}| \Rcondi,\Rcondu),
\end{equation}}
where we have defined
$\Rcondi=\{R_{o_{t'}}:t'<t,o^j_{t'}=o^j_t\}$ as the behaviors on item $o^j_t$ in $R_{o_{<t}}$, which form all the UUCs of $R_{o_t}$ (by time $t$).
$\Rcondu$ is defined symmetrically.


Inspired by NADE~\cite{larochelle2011neural} and CF-NADE~\cite{zheng2016neural},
we model the conditional in \eqn{conditional} with neural networks due to the rich expressive ability.
Specifically, \method{} models the UUCs and IICs of $R_{o_t}$ with hidden representations respectively:
{\small
\begin{align}
    \hv^{U}(\Rcondi) &= f\Big(\sum{}_{t'<t:o^j_{t'}=o^j_t}\Wv^{U}_{:,o^i_{t'},R_{o_{t'}}}+\cv^{U}\Big),\label{eqn:enc_u}\\
    \hv^{I}(\Rcondu) &= f\Big(\sum{}_{t'<t:o^i_{t'}=o^i_t}\Wv^{I}_{:,o^j_{t'},R_{o_{t'}}}+\cv^{I}\Big), \label{eqn:enc_i}
\end{align}}
where $f(\cdot)$ is a nonlinear function, such as $\textrm{tanh}(x)=\frac{\exp(x)-\exp(-x)}{\exp(x)+\exp(-x)}$, $\Wv^{U}\in\RR^{H_U\times N\times K}$ and $\Wv^{I}\in\RR^{H_I\times M\times K}$ are $3$-order tensors,
$\cv^{U}\in\RR^{H_U}$ and $\cv^{I}\in\RR^{H_I}$ are the bias terms.
$H_U$ and $H_I$ are the dimensions of the hidden representations for UUCs and IICs, respectively.
The column $\Wv^{I}_{:,j,k}$ denotes how much ``behaving $k$ on item $j$'' contributes to the hidden representations of the IICs while the column $\Wv^{U}_{:,i,k}$ denotes the contribution of ``user $i$ behaves $k$'' to the hidden representation of the UUCs.

\method{} explains the hidden representations of the UUCs and the IICs by computing the activations:
{\small
\begin{align}
    s^{U}_{o_t^i,k}(\hv^{U}(\Rcondi)) &= \Vv^{U}_{:,o^i_t,k}{}^{\top}\hv^{U}(\Rcondi)+b^{U}_{o^i_t,k},\label{eqn:dec_i}\\
    s^{I}_{o_t^j,k}(\hv^{I}(\Rcondu)) &= \Vv^{I}_{:,o^j_t,k}{}^{\top}\hv^{I}(\Rcondu)+b^{I}_{o^j_t,k},\label{eqn:dec_u}
\end{align}}
 where $\Vv^{U}\in\RR^{H_U\times N\times K}$ and $\Vv^{I}\in\RR^{H_I\times M\times K}$ are $3$-order tensors,
$\bv^{U}\in\RR^{N\times K}$ and $\bv^{I}\in\RR^{M\times K}$ are the bias terms.
The column $\Vv^{I}_{:,j,k}$ is the coefficients that determine how the hidden representation of the IICs affects the activation $s^{I}_{j,k}$ for ``behaving $k$ on item $j$''.
Higher activation $s^{I}_{j,k}$ indicates that the considered IICs suggest higher probability that the user will carry out a behavior $k$ on item $j$.
The activation $s^{U}_{i,k}$ is interpreted similarly.

Finally, to combine the activations of UUCs and IICs of $R_{o_t}$ and produce a probability
distribution, we define the final conditional model as a softmax function of the summation of the two activations:
{\small
\begin{equation}\label{eqn:probability}
    p( R_{o_t}=k| R_{o_{<t}}) =
    \dfrac{\exp\left(s^{U}_{o_t^i,k}+s^{I}_{o_t^j,k}\right)
    }{\sum_{k'=1}^{K}\exp\left(s^{U}_{o_t^i,k'}+s^{I}_{o_t^j,k'}\right)}.
\end{equation}}
\fig{workflow} illustrates the conditional model.

\subsubsection{Orderings in Different Tasks}
\label{sec:autoregression}

We now discuss the effect of different orderings on the model and show what kinds of orderings are considered for two major CF tasks detailedly.

\begin{figure}[tb]
\centering
\begin{subfigure}[b]{.28\linewidth}
\includegraphics[width=0.96\textwidth]{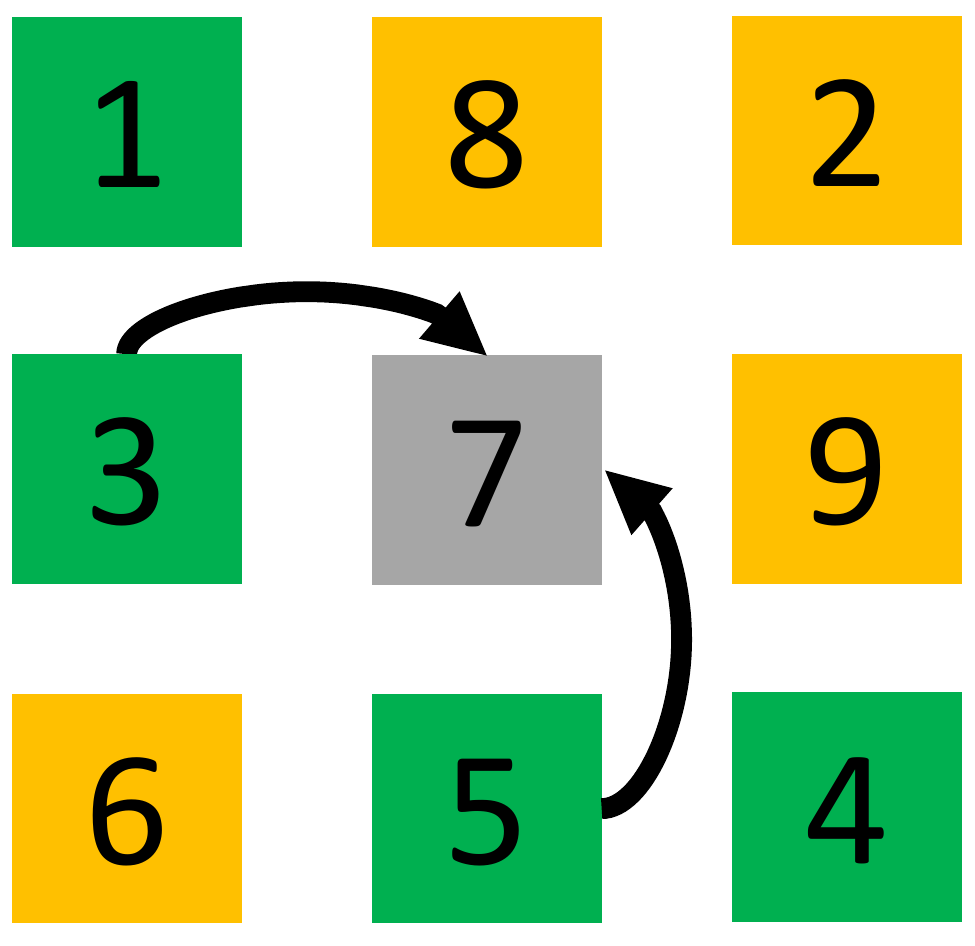}
\caption{}\label{fig:ordera}
\end{subfigure}
\hfill
\begin{subfigure}[b]{.28\linewidth}
\includegraphics[width=0.96\textwidth]{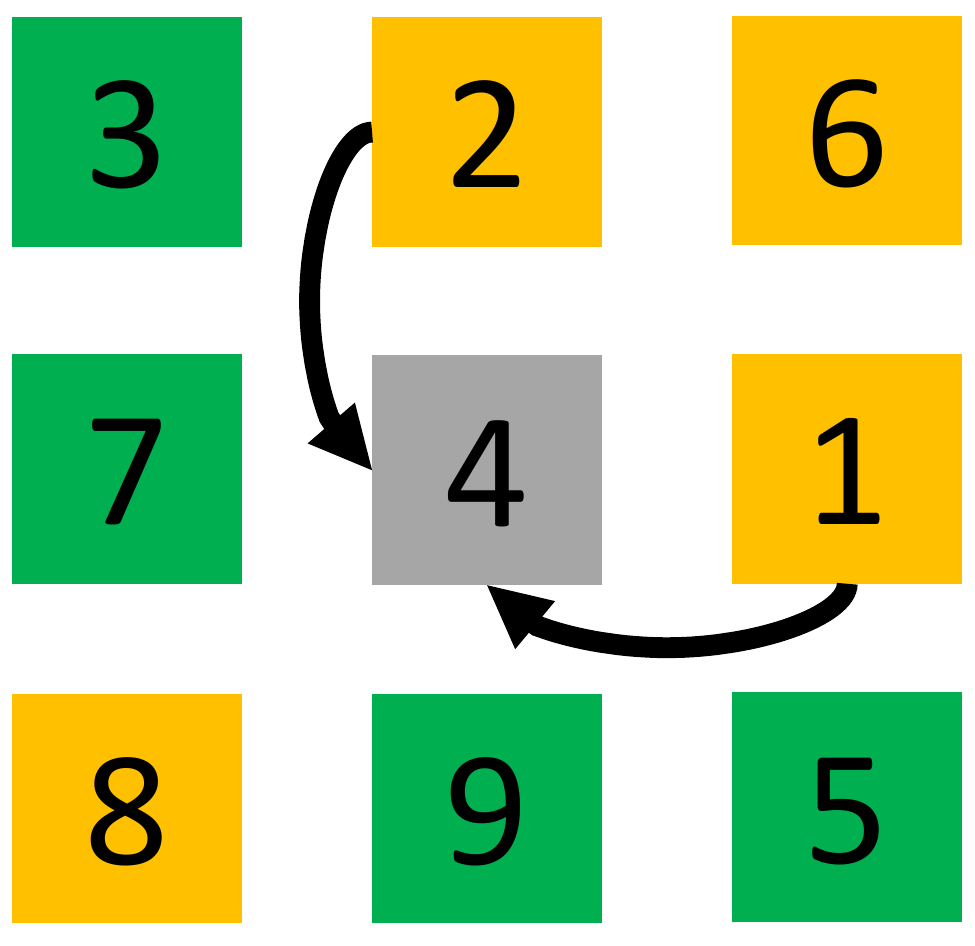}
\caption{}\label{fig:orderb}
\end{subfigure}
\hfill
\begin{subfigure}[b]{.28\linewidth}
\includegraphics[width=0.96\textwidth]{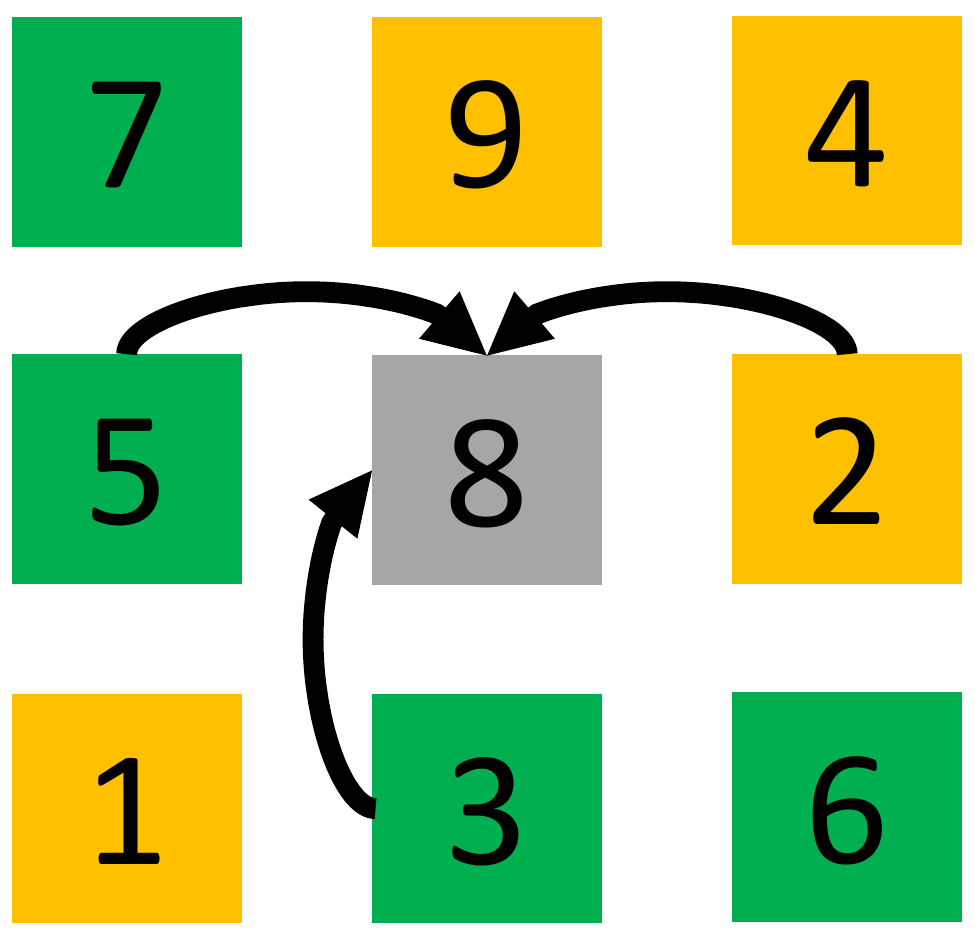}
\caption{}\label{fig:orderc}
\end{subfigure}
\caption{An illustration of how the orderings decide the conditional models. Colors are interpreted same as in \fig{workflow}. (a) - (c) show $3$ conditional models for the central (gray) behavior in an example behavior matrix under 3 different orderings. The numbers $1$ - $9$ indicate the indices of the corresponding behaviors in the orderings. The arrows indicate the dependencies involved in the conditional models. }\label{fig:ordering}
\end{figure}

In fact, the ordering $o$ decides the conditional model for each observed behavior $R_{i',j'}$. Specifically, according to our model (See Eqns.~(\ref{eqn:conditional}) to (\ref{eqn:enc_i})),
the contributions of UUCs and IICs to a given behavior $R_{i',j'}$, i.e. $R^{UUC}_{i',j'}$ and $R^{IIC}_{i',j'}$, depend on where the ordering $o$ places $R_{i',j'}$ and what $o$ places before $R_{i',j'}$.
In general, different orderings result in different conditional models or dependency relations (see \fig{ordering} for an illustration) and any possible conditional models can be induced by some specific orderings. 
Such a property leaves us freedom to control what kind of dependencies we would like the model to exploit in different tasks, as shown below.

CF methods are usually evaluated on rating prediction tasks~\cite{zheng2016neural,sedhain2015autorec}, or more generally, matrix completion tasks, by predicting randomly missing ratings/values.
For matrix completion tasks, taking all UUCs and IICs into consideration leads the model to exploit the underlying correlations to a maximum extent.
Therefore, 
we should consider all possible conditional models for each behavior, i.e., all orderings, in such tasks.
The objective could then be defined as the expected (over all orderings) negative log-likelihood (NLL) of the training set:
{\small
\begin{align}
    \mathcal{L}(\thetav)&=\ep_{o\in\mathfrak{S}_D}-\log p(\data|\thetav,o) \nonumber \\
    &=-\ep_{o\in\mathfrak{S}_D}\sum{}_{d=1}^D\log p(R_{o_d}|R_{o_{<d}},\thetav,o), \label{eqn:obj}
\end{align}}
where $\thetav$ denotes all the model parameters and
$\mathfrak{S}_{D}$ is the set of all the permutations of $\data{}$\footnote{Given the training set $\data{}$, the first $D$ elements of $o$ will be automatically restricted to $\data{}$. As we only evaluate the likelihood of the training set $\data{}$, the number of equivalence orderings are $D!$.}.
Note that taking the expectation over all orderings is equivalent to integrating them out.
Thus the training procedure does not depend on any particular ordering and no manually chosen ordering is needed.

Recent works~\cite{rendle2009bpr,he2017neural} also evaluate CF on top-N recommendation tasks, aiming to suggest a short list of future preferences for each user, which is closer to the goal of real-world recommendation systems. 
In these tasks,
not all IICs are useful. 
For example, people who have just watched \textit{Harry Potter 1} (\textit{HP1}) are very likely to be interested in \textit{Harry Potter 2} (\textit{HP2}), however those who have just watched \textit{HP2} are less likely to have interest in \textit{HP1}, as he/she may have known some spoiler about \textit{HP1}.
To this end, we should expect the model to capture the chronological IICs, which only include the dependencies from later behaviors to previous behaviors of each user, and all UUCs in the meanwhile.
Then, an appropriate objective should be the expected NLL of the training set over all orderings that do not break the chronological order of each user's behaviors.
Note that this can be implemented equivalently by re-defining $R^{IIC}_{i',j'}=\{R_{i',j''}: T(R_{i',j''})<T(R_{i',j'})\}$, where $T(\cdot)$ is the timestamp when the system observes the behavior, and using \eqn{obj} as the objective\footnote{In this case $R^{IIC}_{i',j'}$ is deterministic and only $R^{UUC}_{i',j'}$ depends on the ordering $o$.}. Hence we still need not to choose any ordering manually.

The above examples show that extra desired properties can be incorporated into \method{} for different tasks by considering different orderings in the objective, which indeed benefits from the co-autoregressive assumption.

\subsection{Learning}\label{sec:learning}

The remaining challenge is to optimize the objective in \eqn{obj}.
A common strategy to deal with complicate integrations or large-scale datasets is to adopt stochastic optimization approaches, e.g. stochastic gradient descent (SGD)~\cite{bottou2010large,kingma2014adam}, which require an unbiased estimator of the objective or its gradient.
SGD and its variants have been widely adopted in various areas due to its efficiency,
including
many CF methods~\cite{sedhain2015autorec,zheng2016neural}.
However, unlike in the most existing neural networks based methods~\cite{wu2016collaborative,zheng2016neural}, the users are not modeled independently in \method{}, resulting the objective cannot be estimated stochastically by simply sub-sampling the users.
To tackle this challenge, we derive an unbiased estimator of \eqn{obj} below,
which completes the proposed method.

By exchanging the order of the expectation and the summation in \eqn{obj} and doing some simple math, we get:
{\small
\begin{equation}\label{eqn:sumod}
    \mathcal{L}(\thetav)\!=\!-\!\!\!\!\!\!\sum_{(i',j')\in\data{}}\!\ep_d\ep_{o\in\mathfrak{S}_D|o_d=(i',j')}\!\!\!\log p(R_{i',j'}|R_{o_{<d}},\!\thetav,\!o).
\end{equation}}

According to the definition of the conditional model from Eqns. (\ref{eqn:enc_u}) to (\ref{eqn:probability}),
the log-probability of $R_{i',j'}$ in \eqn{sumod} depends on at most $\Rv_{i',\neg j'}=\Rv_{i',:}\backslash\{R_{i',j'}\}$ (behaviors of user $i'$ except $R_{i',j'}$) and $\Rv_{\neg i', j'}=\Rv_{:,j'}\backslash\{R_{i',j'}\}$ (behaviors on item $j'$ except $R_{i',j'}$).
Specifically, given $o_d=(i',j')$, the log-probability of $R_{i',j'}$ depends on exactly the set $R^{IIC}_{i',j'}=\Rv_{i',\neg j'}\cap R_{o_{<d}}$ and the set $R^{UUC}_{i',j'}=\Rv_{\neg i',j'}\cap R_{o_{<d}}$.
As we treat the ordering $o$ as a random variable uniformly distributed over $\mathfrak{S}_D$, $R^{IIC}_{i',j'}$ and $R^{UUC}_{i',j'}$ are also random.
Moreover, since $R^{IIC}_{i',j'}\cap R^{UUC}_{i',j'}=\varnothing$, they are independent given their sizes $m=|R^{IIC}_{i',j'}|$ and $n=|R^{UUC}_{i',j'}|$, i.e., $R^{IIC}_{i',j'}|m\perp R^{UUC}_{i',j'}|n$.
By expanding the second expectation in \eqn{sumod} based on the above analysis, we have:
{\small
\begin{equation}\label{eqn:summn}
\begin{split}
    \mathcal{L}(\thetav)=&-\sum_{i'=1}^N\sum_{j'=1}^M\ep_{d}\>\ep_{m,n|d}\>\ep_{R^{IIC}_{i',j'}|m}\>\ep_{R^{UUC}_{i',j'}|n}\\
    &\log p(R_{i',j'}|R^{UUC}_{i',j'},R^{IIC}_{i',j'},\thetav)\indicator_{[(i',j')\in\data]},
\end{split}
\end{equation}}
where $m$, $n$, $R^{IIC}_{i',j'}|m$ and $R^{UUC}_{i',j'}|n$ are all random and are decided by the random ordering $o$.
Note the summation over $\data$ is replaced by an equivalent representation using an indicator function $\indicator_{[(i',j')\in\data]}$.
Given $R^{UUC}_{i',j'}$ and $R^{IIC}_{i',j'}$, the log-probability term and the indicator term can be computed easily. From now we omit these two terms for simplicity.


According to symmetry, it is easy to know that $R^{IIC}_{i',j'}|m$ and $R^{UUC}_{i',j'}|n$ are uniformly distributed over all subsets of size $m$ of $\Rv_{\neg i',j'}\cap\data$ and subsets of size $n$ of $\Rv_{i',\neg j'}\cap\data$, respectively.
However, these distributions have different supports since the numbers of the observed behaviors for users (items) are different, which makes the sampling unparallelizable.
Note that the process of drawing $o$ from $\mathfrak{S}_{D}$ can be equivalently simulated by first randomly drawing $\sigma$ from $\mathfrak{S}_{N\times M}$, which can be viewed as an ordering of all the entries of $\Rv$, and then dropping the unobserved entries $\Rv\backslash\data$.
The resulted ordering on $\data{}$ is still uniformly distributed over $\mathfrak{S}_{D}$. Then \eqn{summn} can be written equivalently as:
{\small
\begin{equation}\label{eqn:sumyz}
    \mathcal{L}(\thetav)=-\sum_{i'=1}^N\!\sum_{j'=1}^M\ep_r\>\ep_{y,z|r}\>\ep_{\mathcal{M}\mkern-2mu\subseteq\mkern-2mu[M]\backslash \{j'\}|y}\>\ep_{\mathcal{N}\mkern-2mu\subseteq\mkern-2mu[N]\backslash \{i'\}|z},
\end{equation}}
where $r$ is the index of $R_{i',j'}$ in $\sigma$, $y$ and $z$ are the number of entries in $\Rv_{i',\neg j'}\cap R_{\sigma_{<r}}$ and $\Rv_{\neg i',j'}\cap R_{\sigma_{<r}}$,
respectively. $\mathcal{M}$ is a subset of size $y$ of $[M]\backslash \{j'\}$ and $\mathcal{N}$ is a subset of size $z$ of $[N]\backslash \{i'\}$,
where $[N]$ denotes $\{1,\cdots,N\}$.
$R^{UUC}_{i',j'}$ and $R^{IIC}_{i',j'}$ are therefore $\Rv_{\mathcal{N},j'}\cap\data{}$, $\Rv_{i',\mathcal{M}}\cap\data{}$.

Finally, with some simple math we obtain:
{\small
\begin{equation}\label{eqn:stochastic}
    \mathcal{L}(\thetav)\!=\!-N M\ep_r\>\ep_{y,z|r}\>\ep_{\mathcal{M}\mkern-2mu\subseteq\mkern-2mu[M]|y}\>\ep_{\mathcal{N}\mkern-2mu\subseteq\mkern-2mu[N]|z}\>\ep_{i'\in[N]\backslash\mathcal{N}}\>\ep_{j'\in[M]\backslash\mathcal{M}}.
\end{equation}}
In \eqn{stochastic}, $y$ and $z$ can be computed after sampling $r$ and $\sigma$. $\mathcal{M}$ and $\mathcal{N}$ are sampled by uniformly choosing $y$ and $z$ elements in $[M]$ and $[N]$ without replacement, respectively.
The last two expectations can be estimated unbiasedly by sampling $B_U$ elements from $[N]\backslash\mathcal{N}$ and $B_I$ elements from $[M]\backslash\mathcal{M}$, respectively, where $B_U$ and $B_I$ can be viewed as the minibatch sizes of users and items.
Finally, we get an unbiased estimation of 
the objective $\mathcal{L}(\thetav)$, which can be then adopted in SGD algorithms.

Note that though the training objective involves expectations over multiple orderings (which help exploit the desired UUCs and IICs during training), 
the prediction procedure is simple and deterministic. For an unknown behavior $R_{i^*,j^*}$, the prediction is evaluated with $\hat{R}_{i^*,j^*}=\mathbb{E}_{p(R_{i^*,j^*}=k|\data{})}[k]$ 
with $R^{UUC}_{i^*,j^*}=\Rv_{\neg i^*,j^*}\cap\data$ and $R^{IIC}_{i^*,j^*}=\Rv_{i^*,\neg j^*}\cap\data$, where we have assumed $R_{i^*,j^*}=R_{o_{D+1}}$ and $R_{o_{<D+1}}=\data{}$.


\section{Experiments}\label{sec:exp}

We now present a series of experimental results of the proposed \method{} to demonstrate its effectiveness.
We compare \method{} with other popular CF methods on two major kinds of CF tasks:
rating prediction and top-N recommendation.
The experiments are conducted on two representative 
datasets:
\mlone{}~\cite{harper2016movielens} and \netflix{}~\cite{bennett2007netflix}.
\mlone{} consists of $1,000,209$ ratings of $3,952$ movies (items) rated by $6,040$ users.
\netflix{} consists of $100,480,507$ ratings of $17,770$ movies rated by $480,189$ users.
The ratings in both datasets are $1$-$5$ stars scale, i.e., $K=5$.
For all experiments, we use Adam~\cite{kingma2014adam} to optimize the objectives with an initial learning rate $0.001$.
During training, we anneal the learning rate by factor $0.25$ until no significant improvement can be observed on validation set.
Note that in \eqn{stochastic} the sizes of $[N]\backslash\mathcal{N}$ and $[M]\backslash\mathcal{M}$, i.e. $N-z$ and $M-y$, vary from $1$ to $N-1$ and to $M-1$, respectively. As a consequence, the minibatch sizes of users/items should be set dynamically.
Nevertheless, we choose fixed minibatch sizes of users/items $B_U/B_I$, which only take effect when $M-y>B_I$ or $N-z>B_U$.\label{sec:exp:minibatch}
We adopt weight decay on model parameters to prevent the model from overfitting.
Other hyper parameters and detailed experimental settings will be specified latter for each task.
The codes and more detailed settings can be found at
\url{https://github.com/thu-ml/CF-UIcA}.

\begin{figure}[tb]
\centering
\begin{minipage}[t]{.22\textwidth}
\includegraphics[width=0.99\textwidth]{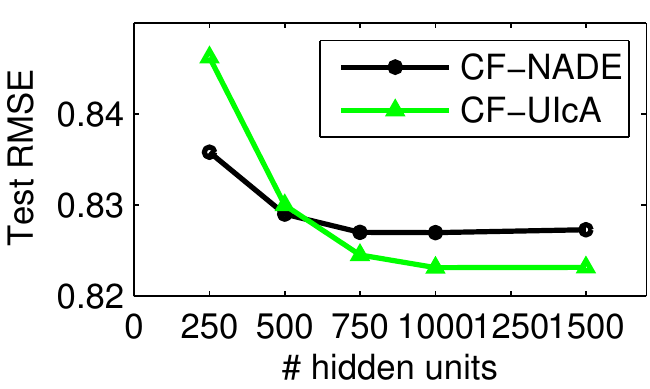}
\caption{The performance on \mlone{} of \method{} and CF-NADE w.r.t. the number of hidden units.}\label{fig:hidden}
\end{minipage}\hfill
\begin{minipage}[t]{.22\textwidth}
\includegraphics[width=0.99\textwidth]{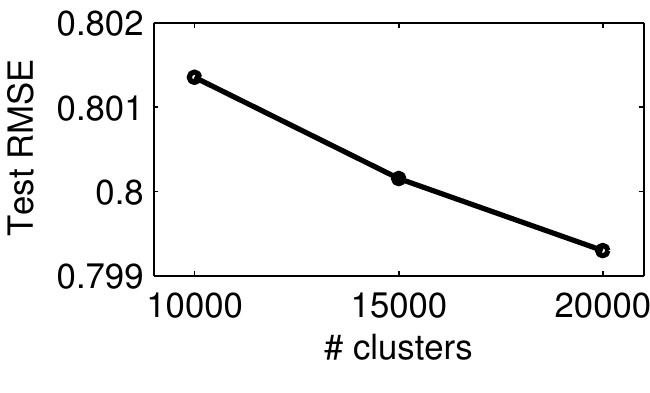}
\caption{The performance on \netflix{} of \method{} w.r.t. the number clusters of users.}\label{fig:cluster}
\end{minipage}
\end{figure}

\subsection{Rating Prediction}\label{sec:exp:explicit}

We use the same experimental settings with LLORMA~\cite{lee2013local}, AutoRec~\cite{sedhain2015autorec} and CF-NADE~\cite{zheng2016neural}.
We randomly select $90\%$ of the ratings in each of the datasets as the training set, leaving the remaining $10\%$ of the ratings as the test set.
Among the ratings in the training set, $5\%$ are hold out for validation.
We compare the predictive performance
with other state-of-the-art methods in terms of the common used \textit{Root Mean Squared Error} (RMSE) $ = 
(\sum{}_{(i,j)\in\data{}_{\textrm{test}}}(\hat{R}_{i,j}-R_{i,j})^2/D_{\textrm{test}})^{1/2},
$
where $\data{}_{\textrm{test}}$ is the test set of $D_{\textrm{test}}$ unknown ratings, $R_{i,j}$ is the true rating and $\hat{R}_{i,j}$ is the prediction.
The reported results are averaged over 10 random splits, with standard deviations less than $0.0002$.

\begin{table}[tb]
\centering
\caption{Test RMSE on \mlone{} and \netflix{}.
All the baseline results are taken from~\protect\citet{zheng2016neural}.
}\label{table:predict}
\begin{tabular}{lcc} \hline
Method                           & \mlone{} & \netflix{} \\
\hline
PMF                              & $0.883$ & - \\
U-RBM                            & $0.881$ & $0.845$  \\
U-AutoRec                        & $0.874$ & -  \\
LLORMA-Global                    & $0.865$ & $0.874$  \\
I-RBM                            & $0.854$ & -  \\
BiasMF                           & $0.845$ & $0.844$  \\
U-CF-NADE-S (2 layers)           & $0.845$ & $0.803$  \\
NNMF                             & $0.843$ & -  \\
LLORMA-Local                     & $0.833$ & $0.834$  \\
I-AutoRec                        & $0.831$ & $0.823$  \\
I-CF-NADE-S (2 layers)           & $0.829$ & -  \\[4pt]
\method{} ($H^U\!\!=\!\!H^I\!\!=\!\!500$) & $\mathbf{0.823}$ & $\mathbf{0.799}$ \\
\hline\end{tabular}
\end{table}

\subsubsection{\mlone{}}\label{sec:exp:ml1m}

For experiments on \mlone{},
we set $B_U/B_I$ to $1,000/1,000$ and the weight decay to $0.0001$.

Since \method{} has a connection with CF-NADE~\cite{zheng2016neural} as mentioned in \secref{basic}, we first present a comparison between \method{} and CF-NADE in \fig{hidden}. CF-NADE models each user with a latent representation, similar with our hidden representation of UUCs or IICs. For fairness, we compare the two methods with the same number of hidden units, where in our \method{} the number of hidden units is $H^U\!\!+\!\!H^I$. Note that in \method{} $H^U$ is not necessarily equal to $H^I$, we nevertheless choose $H^U\!\!=\!\!H^I$ for simplicity. We report the item-based CF-NADE results under best setting as described in \cite{zheng2016neural}.

From \fig{hidden} we observe that for small number of hidden units, e.g. $250$, our method gives a worse result than CF-NADE. This is attributed to that the hidden dimensions allocated to the hidden representation of UUCs and IICs are too small $(H^U\!\!=\!\!H^I\!\!=\!\!125)$ to capture the underlying information. As the number of hidden units increases, we observe \method{} outperforms CF-NADE since \method{} can capture both UUCs and IICs while the item-based CF-NADE can only capture UUCs.
One thing worth mentioning is that the total number of parameters in \method{} is only around $83\%$ of the number of parameters in CF-NADE for \mlone{} when the number of hidden units are same, which implies that \method{} can capture more information than CF-NADE with fewer parameters.

\tabl{predict} (middle column) compares the performance of \method{} with state-of-the-art methods on \mlone{}. The hidden dimensions of \method{} are $H^U=H^I=500$. Our method achieves an RMSE of $0.823$, outperforming all the existing strong baselines.
Note that as RMSE scores have been highly optimized in previous work, our 0.006 RMSE improvement w.r.t. CF-NADE is quite significant, compared to the 0.002 by which CF-NADE improves over AutoRec and 0.002 by which AutoRec improves over LLORMA.

\begin{table}[tb]
\centering
\caption{Test HR@$10$ and NDCG@$10$ of various methods on \mlone{}. The results of baseline methods (except CDAE) are kindly provided by~\protect\citet{he2017neural}.
}\label{table:implicit}
\begin{tabular}{lcc} \hline
Method           & HR@$10$ & NDCG@$10$ \\
\hline
ItemPop          & $0.454$  & $0.254$\\
ItemKNN          & $0.623$  & $0.359$\\
BPR~\cite{rendle2009bpr}	             & $0.690$  & $0.419$\\
eALS~\cite{he2016fast}& $0.704$  & $0.433$\\
NeuMF	         & $0.730$  & $0.447$\\
CDAE             & $0.726$  & $0.448$\\[4pt]
\method{} (Uniform)        & $0.692$ &$0.406$\\
\method{} (Inverse)        & $0.616$ &$0.353$\\
\method{}        & $\mathbf{0.736}$ &$\mathbf{0.492}$\\
\hline\end{tabular}
\end{table}

\subsubsection{\netflix{}}\label{sec:exp:netflix}

The \netflix{} dataset is much bigger than \mlone{}, especially the number of users.
We opt to cluster all the $480,189$ users into $10K$, $15K$ and $20K$ groups, respectively, and make the users in same clusters sharing their corresponding columns in $\Wv^U$ and $\Vv^U$.
To cluster the users, we first run matrix factorization~\cite{libmf} with rank $100$ on the training set.
(Predicting the test set with the learned vectors by MF gives an RMSE of $0.865$.)
Then the clustering process is simply done by running a K-means clustering algorithm on the latent vectors of users learned by MF.
For \method{}, the weight decay is set to $5\times10^{-6}$ as the dataset is sufficiently large.
The minibacth sizes of users and items are set to $B_U=4,000$ and $B_I=1,000$.
The hidden dimensions are $H^U=H^I=500$.


\fig{cluster} shows the performance of \method{} with different number of clusters.
We observe that the performance improves as the number of clusters increases, which can be attributed to that using more clusters empowers the model to capture more variety among users.
Another observation is that the performance can be potentially improved by further increasing the number of clusters.
We do not increase the number of clusters due to the limitation of GPU memory.

\tabl{predict} (right column) summarizes our best result and other state-of-the-art results.
Symbol ``-'' indicates that the authors didn't report the result, probably due to the lack of scalability\footnote{We confirmed with the authors of \cite{zheng2016neural} that I-CF-NADE is not scalable to Netflix. For AutoRec, the authors reported that I-AutoRec is their best model.}.
Our method with $20,000$ clusters of users achieves a state-of-the-art RMSE of $0.799$,
which, together with the results on \mlone{}, proves that our \method{} has the ability to predict users' ratings precisely.

\begin{figure}[tb]
\centering
\includegraphics[width=0.35\textwidth,height=0.22\textwidth]{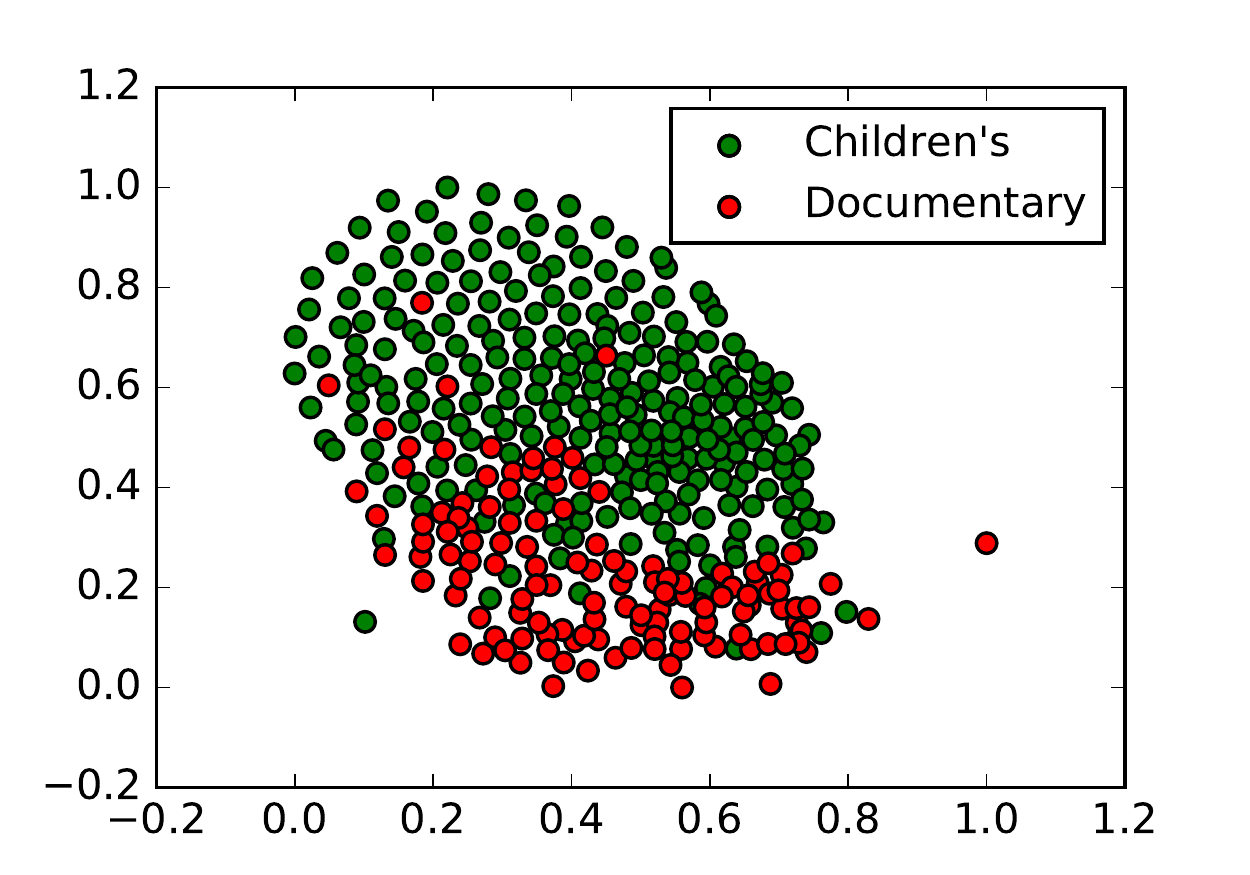}
\caption{t-SNE embedding of the learned vectors for \mlone{}.}\label{fig:tsne}
\end{figure}

\subsection{Top-N Recommendation}\label{sec:exp:implicit}

In most real scenarios, the goal of recommendation systems is to suggest a top-$N$ ranked list of items that are supposed to be appealing for each user.
Moreover, implicit feedback~\cite{zheng2016neural2} has attracted increasing interests because it is usually collected automatically and is thus much more easier to obtain.
We follow the experimental settings of NeuMF~\cite{he2017neural}
to test the recommendation quality of \method{} with implicit feedback.
We transform \mlone{} into implicit data by marking all ratings as 1 indicating that the user has rated the item.
We adopt the \textit{leave-one-out}~\cite{rendle2009bpr,he2017neural} evaluation:
The latest rated item of each user is held out as the test set; The second latest rated item of each user is choosen as the validation set and the remaining data are used as the training set.
At test time, we adopt the common strategy~\cite{koren2008factorization,elkahky2015multi} that randomly samples $100$ items that are not rated by the user,
and ask the algorithm to rank the test item among the 100 sampled items.
We evaluate the quality of the ranked list for the user by computing the \textit{Hit Ratio} (HR) and the \textit{Normalized Discounted Cumulative Gain} (NDCG)~\cite{he2015trirank}.
Specifically,
{\small
\begin{equation}
    \textrm{HR}\!=\!\frac{\textrm{\#hits}}{\textrm{\#users}},~~~~\textrm{NDCG}\!=\!\frac{1}{\textrm{\#users}}\!\sum_{i=1}^{\textrm{\#hits}}\frac{1}{\log_2(p_i\!+\!1)},
\end{equation}}
where \#hits is the number of users whose test item appears in the recommended list and $p_i$ is the position of the test item in the list for the $i$-th hit.
For both metrics, the ranked list is truncated at $10$.

Since the model is always asked to make predictions of latest behaviors based on former behaviors, we train the model under the expectation over orderings that maintain the chronological order of each user's behaviors, as anylized in \secref{autoregression}.
An important difficulty of CF with implicit feedback is that only positive signals are observed.
To handle the absence of negative signals, we follow the common strategy~\cite{pan2008one,he2017neural} that randomly samples negative instances from unobserved entries dynamically during the training procedure.

The minibatch sizes of users and items are set to $200$. The hidden dimensions are $H^U=H^I=256$ and the weight decay is $1\times10^{-5}$.
the results are averaged over $5$ runs with
different random seeds, with standard deviations less than $0.0005$.
\tabl{implicit} compares the results in HR@$10$ and NDCG@$10$ with state-of-the-art methods for top-N recommendation with implicit feedback on \mlone{}. The baseline results are provided by \citet{he2017neural}, except the result of CDAE \cite{wu2016collaborative}, which is evaluated with our implementation.
We can see that \method{} achieves the best performance under both measures. Importantly, our method gives an NDCG@$10$ $0.492$, which outperforms the state-of-the-art method NeuMF by a large margin $0.045$ (relative improvement $10.1\%$).
To demonstrate the significance of the co-autoregression, we train another two \method{} models under the expectation over: (Uniform Case) all possible orderings, which cover all UUCs and IICs; and (Inverse Case) all orderings that reverse the chronological order of each user's behaviors. The results are shown in \tabl{implicit}. We can observe that the orderings significantly affect the performance. Compared to the result (NDCG@$10$ $0.406$) of Ignore case where all UUCs and IICs are captured, our best result brings a $0.09$ improvment, demonstrating the effectiveness of the orderings and the power of the co-autoregression.

\begin{table}[tb]
\centering
\caption{Average running time for each minibatch of \method{} on different datasets.}\label{table:time}
\begin{tabular}{lccc} \hline
Dataset               &  $B_U/B_I$ & $H^U/H^I$ & Time \\
\hline
ML 1M     & $1,000/1,000$ & $500/500$ & $0.77$s \\
\netflix{}& $4,000/1,000$ & $500/500$ & $3.4$s \\
\hline\end{tabular}
\end{table}

\begin{table}[tb]
\centering
\caption{Average test time of different methods and tasks on \mlone{}.}\label{table:testtime}
\begin{tabular}{lccc} \hline
Method               &  Task & Test Time \\
\hline
CF-NADE  & Rating Prediction & $0.68$s \\
\method{}& Rating Prediction & $0.80$s \\
\method{}& Top-N Recommendation & $0.73$s \\
\hline\end{tabular}
\end{table}

\subsection{Visualization}\label{sec:exp:visual}

In \mlone{}, each movie is marked with one or more genres.
There are totally 18 different genres including \textit{Action}, \textit{Children's}, \textit{Drama}, etc.
We visualize the learned weight matrices $\Vv^I$ in \secref{exp:explicit}.
Specifically, once the model is learned, $\Vv^I_{:,j,:}$ can be viewed as $H^I\times K$ dimensional vectors associated with item $j$. We apply t-SNE~\cite{maaten2008visualizing} to embed these vectors into a 2-dimensional plane.
\fig{tsne} shows the t-SNE embedding of two most exclusive genres: \textit{Children's} and \textit{Documentary}.
We can observe the learned vectors are distributed semantically in the gross. 

\subsection{Running Time and Memory}\label{sec:exp:time}

We analyze the running time and memory cost of the proposed method.
All the experiments are conducted on a single Nvidia TITAN X GPU with Theano~\cite{2016arXiv160502688short} codes.
As explained in \secref{exp:minibatch}, the minibatch sizes of users and items are not deterministic during training and thus there is no standard way to train \method{} epoch by epoch. We report the average training time for each minibatch in \tabl{time}.
As for testing time, we compare \method{} with other state-of-the-art methods in \tabl{testtime}. 

The running memory cost of \method{} is mainly for saving the 3-dimensional weight tensors. Specifically, the memory complexity of \method{} is $O((NH^U+MH^I)K)$.
In our experiments we always let $H^U=H^I=H$, resulting the memory cost is proportional to $(N+M)HK$.

\section{Conclusion}\label{sec:conclusion}

We propose \method{}, a neural co-autoregressive model for collaborative filtering, with a scalable stochastic learning algorithm. \method{}  performs autoregression in both users and items domains, making it able to capture both correlations between users and items explicitly and simultaneously. 
Experiments show that our method achieves state-of-the-art results, and is able to learn semantic information by visualization, verifying that the autoregression provides a principled way to incorporate the correlations.

\section*{Acknowledgments}
This work is supported by the National NSF of China (Nos. 61620106010, 61621136008, 61332007), the MIIT Grant of  Int. Man. Comp. Stan (No. 2016ZXFB00001) and the NVIDIA NVAIL Program.

\small
\bibliographystyle{aaai}
\bibliography{bibtex}

\end{document}